\documentclass[journal,transmag]{IEEEtran}
\usepackage{epsfig}
\usepackage{graphicx}
\usepackage{amsmath}
\usepackage{amssymb}
\usepackage{booktabs}
\usepackage{multirow}
\usepackage{tabularx}
\usepackage{booktabs}
\usepackage{makecell}
\usepackage{url}
\usepackage{subcaption}
\usepackage{bm}
\usepackage{threeparttable}
\usepackage{comment}
\ifCLASSINFOpdf
\else
\fi
\begin{document}
%\title{Shortcut Learning of discriminator: High-Frequency Bias is not\\ Beneficial for GANs training}
\title{Exploring The Effect of High-frequency Components in GANs Training}


\author{\IEEEauthorblockN{Ziqiang Li\IEEEauthorrefmark{1},
%Chaoyue Wang\IEEEauthorrefmark{2},
%Jing Zhang\IEEEauthorrefmark{2},
Pengfei Xia\IEEEauthorrefmark{1},
Xue Rui\IEEEauthorrefmark{2}, and
Bin Li\IEEEauthorrefmark{1},~\IEEEmembership{Member,~IEEE}}
\IEEEauthorblockA{\IEEEauthorrefmark{1}School of Information Science and Technology, University of Science and Technology of China, Hei Fei, China}
%\IEEEauthorblockA{\IEEEauthorrefmark{2}School of Computer Science, University of Sydney, Sydney, Australia}
\IEEEauthorblockA{\IEEEauthorrefmark{2}State Key Laboratory of Fire Science, University of Science and Technology of China, Hei Fei, China}
% <-this % stops an unwanted space
\thanks{Manuscript received XX XX, XXXX; revised XX XX, XXXX. Ziqiang Li, Pengfei Xia, and Xue Rui are with the University of Science and Technology of China, Anhui, China. (E-mail: \{iceli, xpengfei, ruixue27\}@mail.ustc.edu.cn). Corresponding author: Bin Li (E-mail: binli@ustc.edu.cn).}}

%{Shell \MakeLowercase{\textit{et al.}}: Bare Demo of IEEEtran.cls for IEEE Transactions on Magnetics Journals}



% for Transactions on Magnetics papers, we must declare the abstract and
% index terms PRIOR to the title within the \IEEEtitleabstractindextext
% IEEEtran command as these need to go into the title area created by
\maketitle
\begin{abstract}
	%Recent studies demonstrate that neural networks with supervised tasks are prone to rely on some shortcut features for decisions. Consequently, these networks could not generalize to out-of-distribution (o.o.d) sets. Contrary to the supervised task, GAN is an unsupervised task, which is more susceptible to shortcut features. The discriminator focusing on shortcut features cannot obtain generalized features for authenticity, which seriously affects the training of generator. In this work, we demonstrate that the discriminator in GANs is sensitive to high-frequency components that cannot be distinguished by humans and the high-frequency components of images can be considered as shortcut features of the discriminator, which are not conducive to the training of GANs. Based on these, we propose two preprocessing methods mitigating the shortcut learning in GANs training: High-Frequency Confusion (HFC) and High-Frequency Filter (HFF). The proposed methods are general and can be easily applied to most existing GANs frameworks with a fraction of the cost. The advanced performance of the proposed method is verified on multiple loss functions, network architectures, and datasets.
	Generative Adversarial Networks (GANs) have the ability to generate images that are visually indistinguishable from real images. However, recent studies have revealed that generated and real images share significant differences in the frequency domain. In this paper, we explore the effect of high-frequency components in GANs training. According to our observation, during the training of most GANs, severe high-frequency differences make the discriminator focus on high-frequency components excessively, which hinders the generator from fitting the low-frequency components that are important for learning images' content. Then, we propose two simple yet effective frequency operations for eliminating the side effects caused by high-frequency differences in GANs training: High-Frequency Confusion (HFC) and High-Frequency Filter (HFF). The proposed operations are general and can be applied to most existing GANs with a fraction of the cost. The advanced performance of the proposed operations is verified on multiple loss functions, network architectures, and datasets. Specifically, the proposed HFF achieves significant improvements of $42.5\%$ FID on CelebA (128*128) unconditional generation based on SNGAN, $30.2\%$ FID on CelebA unconditional generation based on SSGAN, and $69.3\%$ FID on CelebA unconditional generation based on InfoMAXGAN.
\end{abstract}

%%%%%%%%% BODY TEXT
\section{Introduction}
%Shortcut learning \cite{geirhos2020shortcut} of Deep Neural Networks (DNN) is defined as the decision rules that perform well on standard benchmarks but fail to transfer to more challenging testing conditions. Shortcut learning is discovered through cases of unintended generalization, revealing a mismatch between human-intended and model-learned solutions. Recently, many studies are proposed to mitigate shortcut learning in different domains, such as classification \cite{hermann2020shapes,dagaev2021too}, Natural Language Understanding \cite{du2021towards}, and Visual Question Answering \cite{dancette2021beyond}. In this paper, we analyze the shortcut learning of frequency in GANs training for the first time.

\begin{figure}
	\begin{subfigure}[b]{0.45\textwidth}
		\centering
		\includegraphics[width=0.85\textwidth]{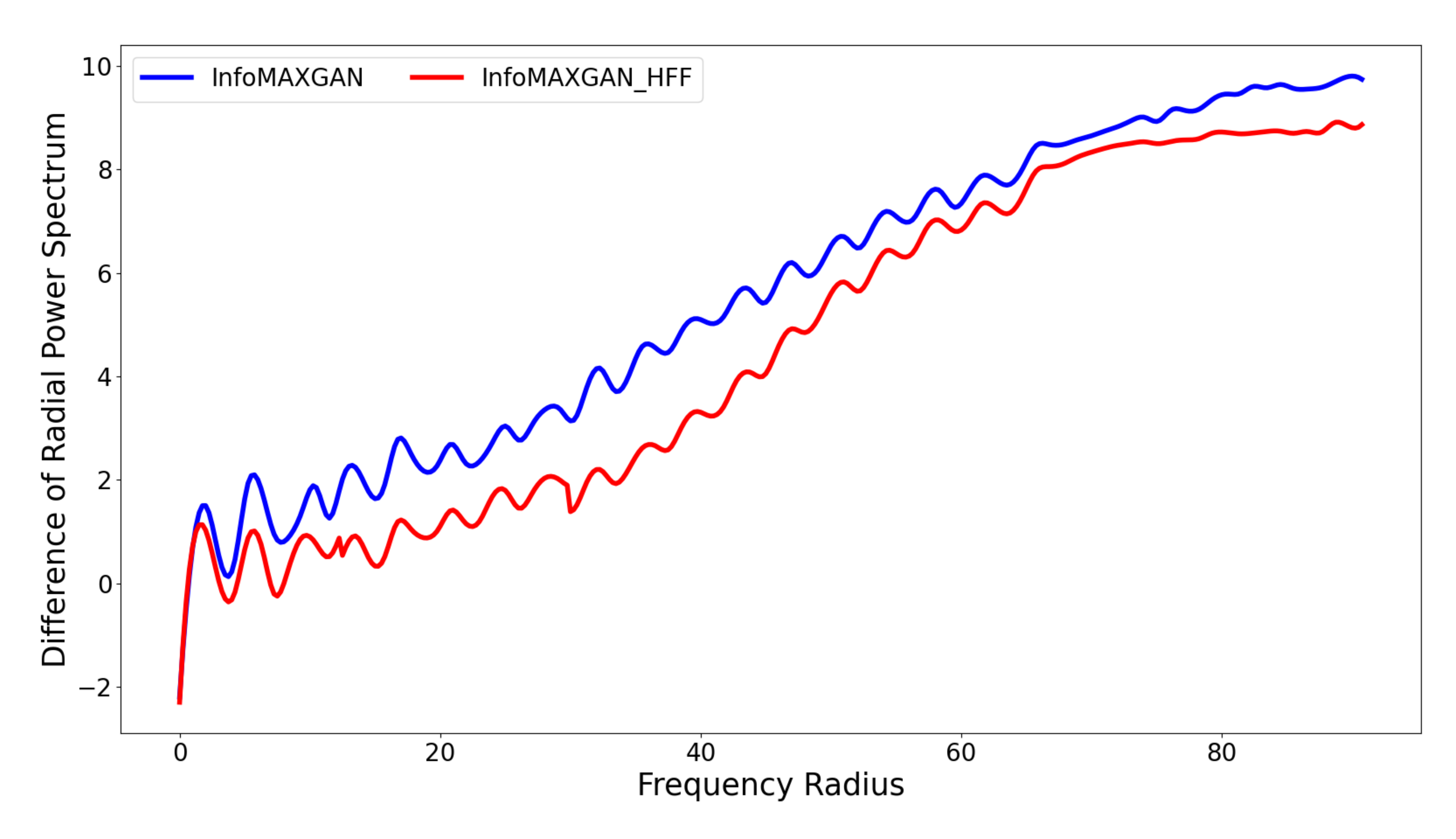}
		\caption{The difference of mean after azimuthal integration over the Radial Power Spectrum (see Section 3.1) of the images, where red line represents the difference between real and fake images generated by InfoMAXGAN with HFF, amd blue line represents the difference between real and fake images generated by InfoMAXGAN. All results are calculated on a randomly selected sample of 1000.}
		\label{figure:frequency of confusion}
	\end{subfigure}
	\begin{subfigure}[b]{0.45\textwidth}
		\centering
		\includegraphics[width=0.85\textwidth]{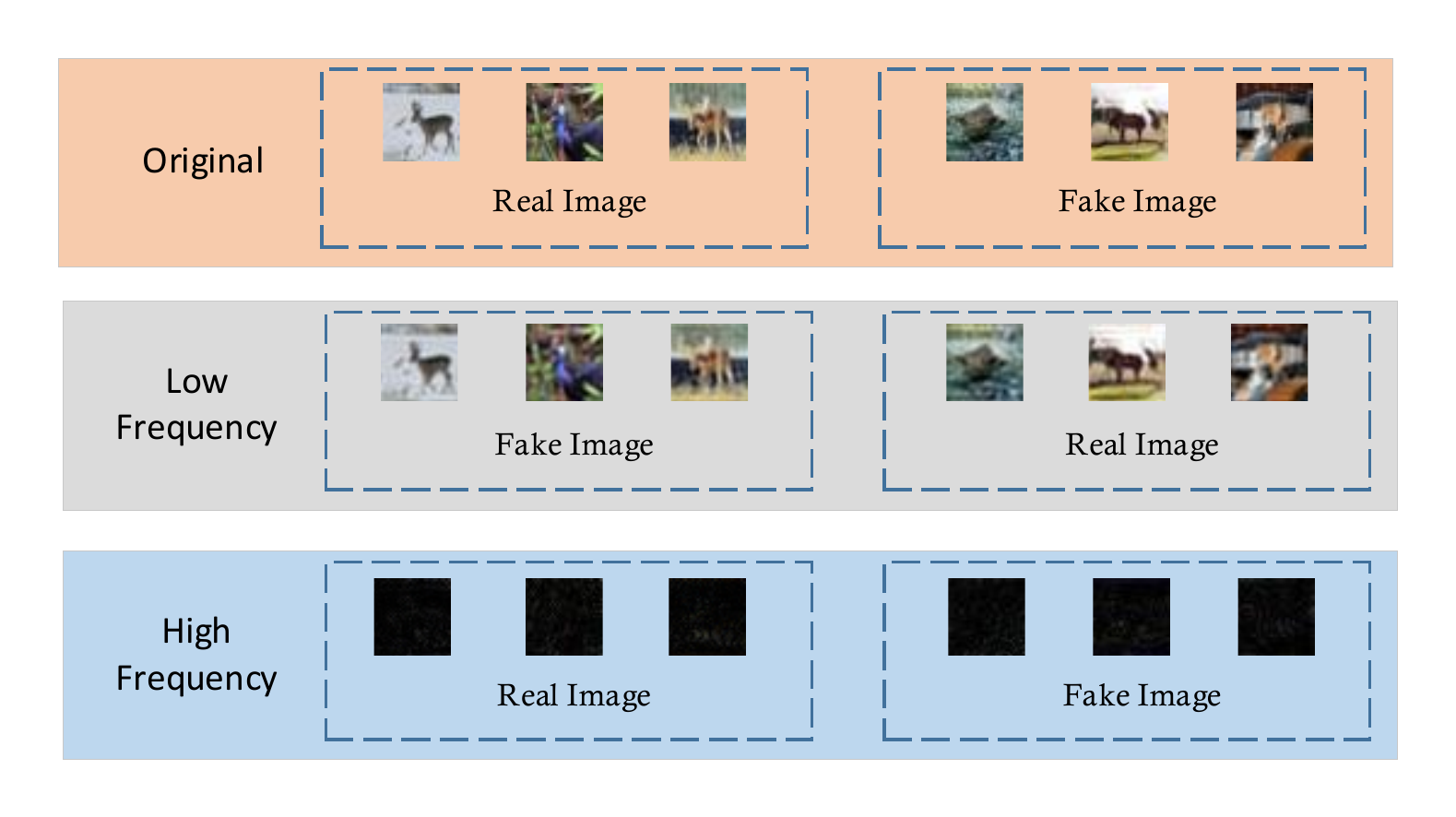}
		\caption{Some examples of predicting defect of discriminator in the frequency domain. The trained discriminator obtains the same predictions between original (top row) and high-frequency components (bottom row). However, the predictions of low-frequency components (middle row) is different from that of original images (the filter Radius for separating different frequency components is set to 10).}
		\label{figure:shortcut}
	\end{subfigure}
	\centering
	\caption{(a): The frequency biases of InfoMAXGAN and InfoMAXGAN with HFF on the CelebA dataset. (b): Discriminator prediction of SNGAN \cite{miyato2018spectral} on the CIFAR-10 dataset.}
	\label{figure:abstract}
\end{figure}

Generative Adversarial Networks (GANs) \cite{goodfellow2014generative} have been widely used in the field of computer vision and image processing: attribute editing \cite{tao2019resattr,shen2020interpreting}, generation of photo-realistic images \cite{karras2019style,karras2020analyzing},  image inpainting \cite{gu2020image,shin2020pepsi++}. Although the remarkable advancements in community, recent works \cite{khayatkhoei2020spatial,dzanic2019fourier,durall2020watch,chen2021ssd,zhang2019detecting,frank2020leveraging} have demonstrated that generated images are significantly different from real images in the frequency domain, especially in the high frequency. %In this paper, we refer to such differences as the frequency bias. 
As shown in Figure \ref{figure:frequency of confusion}, the frequency bias between real and generated images increases with the frequency.
There are two main hypotheses for the artifacts with generated images in the frequency domain: some studies \cite{frank2020leveraging,durall2020watch,zhang2019detecting} suggest that this difference is mainly due to the employed of upsampling operations; and some studies \cite{khayatkhoei2020spatial,dzanic2019fourier} suggest that the linear dependencies in the Conv filter’s spectrum cause spectral limitations.

%Furthermore, the spectral characteristics of neural networks have also attracted the attention of scholars. Wang et al. \cite{wang2020high} discuss the relationship between the frequency spectrum of the images and the generalization of convolutional neural networks (CNN). They think that high-frequency components, which are imperceptible to humans, can improve the performance and reduce the robustness of the classifier. 
%this paper proposes a confusion: \textbf{What is the effect of high frequency components on the training of GANs}.

GANs training is a dynamic game. A good discriminator is a prerequisite for training the generator. Existing discriminators directly use all information of the generated and real images to make the `true/fake' decision, which could introduce some shortcut during GANs training. Figure \ref{figure:shortcut} illustrates several examples of predicting defect of the discriminator in the frequency domain, which indicates that the discriminator is sensitive to the high-frequency components and relies on the high-frequency information as the shortcut to make their decision. In this study, we comprehensively explore and analyze the impact of high-frequency components during the training of GANs. 

High-frequency components of images contain limited information \cite{wang2020high} and are difficult to be fitted \cite{khayatkhoei2020spatial}. Too much focus on high-frequency biases degrades the fit of generator to low-frequency components because the generator may be optimized to fool the discriminators by perturbing high-frequency components. Based on this, we propose two frequency operations (HFC and HFF) for removing the high-frequency biases during the training of GANs, which alleviates the sensitivity of the discriminator to high-frequency biases and focuses more on low-frequency components. Among them, High-Frequency Confusion (HFC) replaces the high-frequency components of the generated images with the high-frequency components of real images, while High-Frequency Filter (HFF) filters out the high-frequency components of both real and generated images. Figure \ref{figure:frequency of confusion} shows the frequency bias between real and fake images generated by InfoMAXGAN and InfoMAXGAN with HFF. The result demonstrates that the proposed method can effectively mitigate the low-frequency bias without compromising the high-frequency fit. Although frequency biases, especially at high frequencies, are not completely eliminated, the proposed methods are general and can be applied to most existing GANs frameworks with a fraction of the cost. We also hope this paper can inspire more works in this area to better solve the frequency bias problem. 
%We hope that more work is presented in this area.

%To mitigate the shortcut learning of frequency in GANs, we propose two data preprocessing methods (HFC and HFF). The proposed methods eliminate the high-frequency bias between real and generated images during the GANs training. Specifically, HFC replaces the high-frequency components of the generated images with the high-frequency components of the real images, while HFF uses a high-frequency filter to filter out the high-frequency components of both real and generated images. The radial power spectrum after preprocessing with HFC and HFF on the CIFAR-10 dataset can be observed from the middle part and bottom part of Figure \ref{figure:frequency of confusion}, respectively.

%Inspired by CNN can exploit the high-frequency components that are not perceivable to humans, this paper analyzes and discusses the impact of high-frequency components on the training of GANs. The experiments indicate that \textbf{the discriminator is sensitive to the high-frequency components that are imperceptible to the human eyes, and these high-frequency components are not conducive to the training of GANs.}

The main contributions of this paper can be summarized as follows:
\begin{itemize}
	\item This paper aims for the first pilot study to analyze the frequency response in GANs. According to experiments analysis, we demonstrate that discriminator is sensitive to the high-frequency components of images and the high-frequency biases between real and fake images affect the fit of low-frequency components during the training of generators.
	\item To balance the high-frequency biases between the generated and the real images during the training of GANs, two simple yet effective frequency operations are proposed in this paper. We confirm that the proposed methods can improve the fit of generator to the low-frequency components effectively.
	\item Comprehensive experiments valid the superiority of our proposed operations. Besides low-frequency components, the proposed operations improve the quality of overall generation on many loss functions, architectures, and datasets with little cost. 
	%\item Through lots of experiments, we obtain the experimental conclusions about the selection of filter Radius (R).
\end{itemize}
\section{Related Works}
\subsection{Generative Adversarial Networks}
Generative Adversarial Networks (GANs) act a two-player adversarial game, where the generator $G(z)$ is a distribution mapping function that transforms low-dimensional latent distribution $p_z$ to target distribution $p_g$. And the discriminator $D(x)$ evaluates the distance between generated distribution $p_g$ and real distribution $p_r$. The generator and discriminator minimize and maximize the zero-sum loss function, respectively. This minimax game can be expressed as:
\begin{equation} 
\begin{split}
\mathop{\min}\limits_{\phi}\mathop{\max}\limits_{\theta} f(\phi,\theta)
=&\mathbb{E}_{x\sim p_{r}}[g_1(D_\theta(x))]\\
+&\mathbb{E}_{z\sim p_z}[g_2(D_\theta(G_\phi(z)))],
\label{Eq:2}
\end{split}
\end{equation}
where $\phi$ and $\theta$ are parameters of the generator $G$ and discriminator $D$, respectively. Specifically, vanilla GAN \cite{goodfellow2014generative} can be described as $g_1(t)=g_2(-t)=-\log(1+e^{-t})$; $f$-GAN \cite{nowozin2016f} can be written as $g_1(t)=-e^{-t}, g_2(t)=1-t$; Morever, Geometric GAN \cite{lim2017geometric} and WGAN \cite{arjovsky2017wasserstein} are described as $g_1(t)=g_2(-t)=-\mathop{\max}(0,1-t)$ and $g_1(t)=g_2(-t)=t$, respectively. 

Recently, the quality of the generated images has been improved dramatically. Karras \textit{et al}. \cite{karras2017progressive} describe a new training methodology in GANs, which grows both the generator and discriminator progressively. The progressive training method makes the quality of high-resolution image synthesis improved unprecedentedly. Also, Karras \textit{et al}. propose StyleGAN \cite{karras2019style} and StyleGAN++ \cite{karras2020analyzing}, which use style-based method to the generation of specific content. Furthermore, some regularization and normalization methods are used to stabilize GANs training \cite{li2020regularization}, WGAN-GP \cite{gulrajani2017improved} and SNGAN \cite{miyato2018spectral} apply 1-Lipschitz continuity using gradient penalty and spectral normalization, respectively; GAN-DAT \cite{li2020direct} minimizes Lipschitz constant to avoid the discriminator being attacked by the adversarial examples; SS-GAN \cite{chen2019self} uses self-supervised techniques to avoid discriminator forgetting; Similarly, InfoMAXGAN \cite{lee2020infomax} improves the image generation via information maximization and contrastive learning, which is an unsupervised method for mitigating catastrophic forgetting. 

In this paper, two frequency operations are used in three loss functions: vanilla GAN loss, LSGAN loss, Geometric loss (Hinge loss) and three frameworks: SNGAN, SSGAN, and InfoMAXGAN. Experiments demonstrate the consistent improvement of our methods.

\subsection{Frequency Principle for CNNs}
Recently, frequency analysis on Convolution Neural Networks (CNNs) has attracted more and more attention. According to the different analysis objects, related studies can be divided into two categories: Input (Image) Frequency and Response Frequency. Many works have shown that there is a principle in both categories of frequency: \textbf{DNNs often fit target functions from low to high frequencies during the training}, which is called \textit{Frequency Principle (F-Principle)} in \cite{xu2019training,xu2019frequency} or \textit{Spectral Bias} in \cite{rahaman2019spectral}. Input (Image) Frequency corresponds to the rate of change of intensity across neighboring pixels of input images, where sharp textures contribute to the high frequency of the image. Wang \textit{et al}. \cite{wang2020high} demonstrate the F-Principle of Input Frequency and point that the high-frequency components of the image improve the performance of the model, but reduce its generalization. 

Furthermore, Response Frequency is the frequency of general Input-Output mapping $f$, which measures the rate of change of outputs with respect to inputs. If $f$ possesses significant high frequencies, then a small change of inputs in the image might induce a large change of the output (e.g. adversarial example in \cite{goodfellow2014explaining}). F-Principle of Response Frequency is first proposed by Xu \textit{et al}. \cite{xu2019training} through regression
problems on synthetic data and real data with principal component analysis. Similarly, Rahaman \textit{et al.} \cite{rahaman2019spectral}  find empirical evidence of a spectral bias: lower frequencies are learned first. They also show that lower frequencies are more robust to random perturbations of the network parameters. Subsequently, Xu \textit{et al}. \cite{xu2018understanding} propose a theoretical analysis framework for one hidden layer neural network with 1-d input, which illustrates activation functions (including Tanh and Relu) in the Fourier domain decays as frequency increases. Furthermore, Xu \textit{et al}. \cite{xu2019frequency} also propose a Gaussian filtering method that can directly verify the F-Principle in high-dimensional datasets for both regression and classification problems. Certainly, there are still a lot of works \cite{xu2018frequency,zhang2019explicitizing,basri2019convergence,cao2019towards,tancik2020fourier} to be done in the frequency domain to interpret the success and failure of CNNs. 

\subsection{Frequency Bias for GANs}
Frequency analysis of GANs is also popular, from which we are mainly concerned with the Input (Image) Frequency. The frequency bias of GANs mainly refers to the bias between the generated images and the real images in the frequency domain, which can be observed in Figure \ref{figure:frequency of confusion}. The same phenomenon has also been found in other works  \cite{frank2020leveraging,zhang2019detecting,dzanic2019fourier,chen2021ssd,durall2020watch,khayatkhoei2020spatial}. All of them demonstrate that this phenomenon is pervasive and hard to avoid, even if some generated images are flawless from human perception. In addition to two main hypotheses (i.e. the employment of upsampling operations and linear dependencies in the Conv filter’s spectrum) introduced in the Introduction, Chen \textit{et al}. \cite{chen2021ssd} reveal that downsampling layers cause the high frequencies missing in the discriminator. This issue may make the generator lacking the gradient information to model high-frequency content, resulting in a significant spectrum discrepancy between generated images and real images. To solve it, they propose the SSD-GAN that boosts a frequency-aware classifier into the discriminator to measure the realness of images in both spatial and spectral domains.

Although the proposed SSD-GAN \cite{chen2021ssd} improves the performance of GANs by increasing the discrimination of discriminator in spectral domain, our proposed HFF and HFC are orthogonal to it and the performance of SSD-GAN can also be further improved. The study that most closely resembles our paper is proposed by Yamaguchi \textit{et al}. \cite{yamaguchi2021f}. Both methods have filtered out unnecessary high-frequency components of the images during the discriminator training. In addition, our method\footnote{The Preprint of our work was published to arXiv (\url{https://arxiv.org/abs/2103.11093}) on March 20, 2021. However, the  Preprint of \cite{yamaguchi2021f} was published to arXiv (\url{https://arxiv.org/pdf/2106.02343}) on June 4, 2021.} analyze the response of discriminator to different-frequency components.

\section{Methodology}
In this section, we first demonstrate frequency decomposition and mixing of images. Then we analyze the discriminator's response for different frequency components. Based on these, we propose two frequency operations to reduce the focus on high-frequency bias between real and fake images during the training of GANs, thereby enhancing the fit of low-frequency components.
\subsection{Frequency Decomposition and Mixing of Images}
For decomposing and mixing different components of images in frequency domain, we first show the DFT $\mathcal{F}$ of 2D image data $I$ with size $N\times N$:
\begin{equation} 
\begin{split}
\mathcal{F}(I)(k, l)&=\sum_{m=0}^{N-1} \sum_{n=0}^{N-1}I(m, n)\cdot e^{-2 \pi j}\big(\frac{ k}{N}m+\frac{l}{N}n\big), \\
\textit{for} \quad k&=0, \ldots, N-1, \quad l=0, \ldots, N-1.
\label{Eq:2}
\end{split}
\end{equation}
And the Power Spectrum (PS) of DFT is defineds as $PS(I)(k,l)=20\cdot\log\left|\mathcal{F}(I)(k, l)\right|$.
At the last, the Radial Power Spectrum (RPS) of DFT via azimuthal integration can be shown as:
\begin{equation} 
\begin{split}
RPS\left(R\right)&=\int_{0}^{2 \pi} PS(I)\left(R \cdot \cos (\phi), R \cdot \sin (\phi)\right) \mathrm{~d} \phi,\\
\textit{for} \quad R&=0, \ldots, N/\sqrt 2,
\label{Eq:2}
\end{split}
\end{equation}
where $R$ is the radius of the frequency. Figure \ref{figure:frequency of confusion} illustrates the RPS of real and fake images on the CelebA dataset. 

\begin{figure}
	\includegraphics[width=0.5\textwidth]{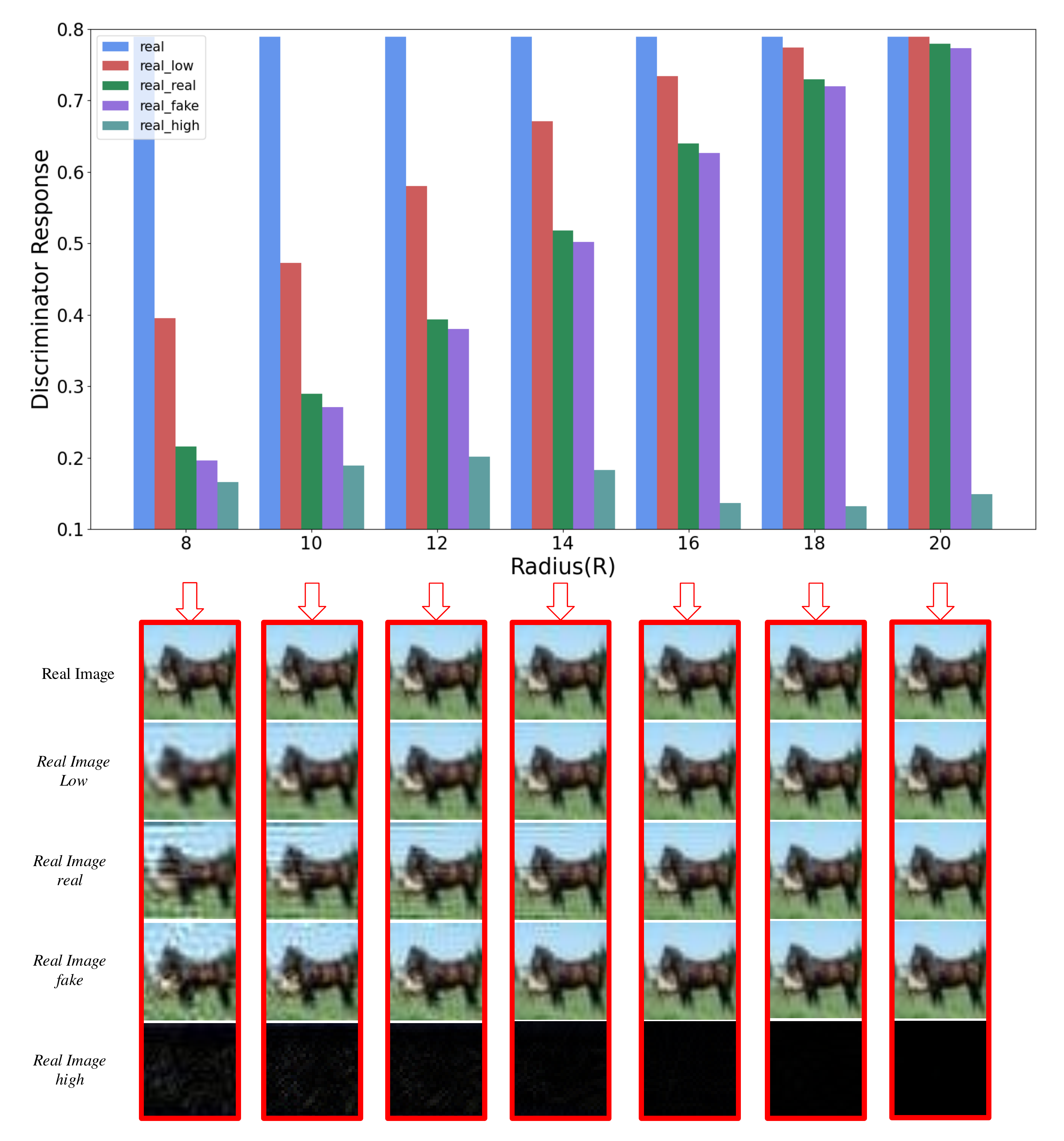}
	
	\centering
	\caption{The response of the discriminator to the real images on CIFAR-10 dataset. \textbf{Top:} the response of the trained discriminator for the real images (The model is SNGAN trained by Hinge Loss. The output of the discriminator goes through the Sigmoid function. All results are calculated on the whole dataset). \textbf{Bottom:} Visualization of different frequency components in the image domain. The image is sampled from the training set.}
	\label{figure:response of the real images}
\end{figure}
\begin{figure}
	\includegraphics[width=0.5\textwidth]{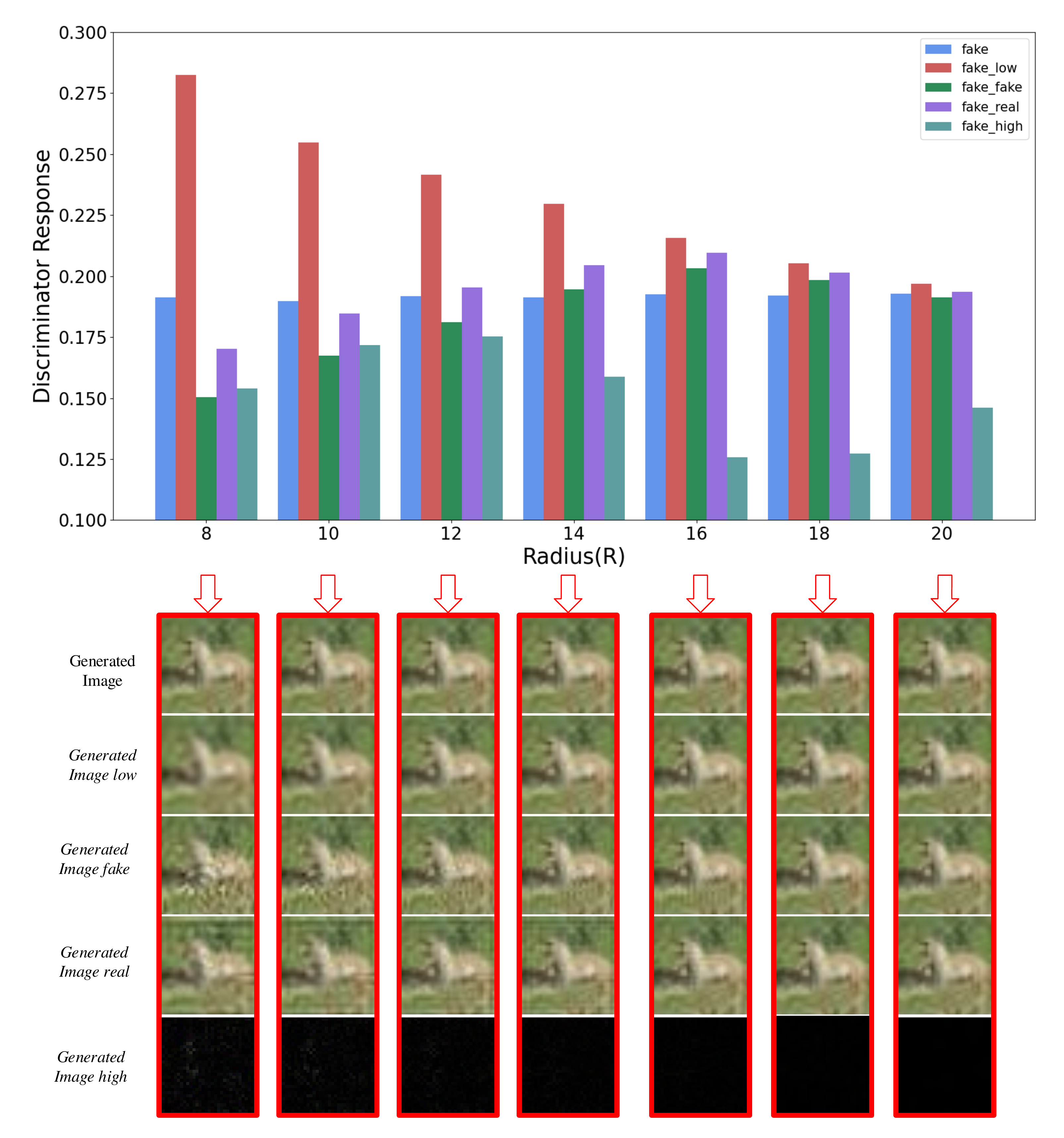}
	
	\centering
	\caption{The response of the discriminator to the generated images on CIFAR-10 dataset. \textbf{Top:} the response of the trained discriminator for the generated images (The model is SNGAN trained by Hinge Loss. The output of the discriminator goes through the Sigmoid function.). \textbf{Bottom:} Visualization of different frequency components in a generated image.}
	\label{figure:response of the fake images}
\end{figure}

In adition, we also introduce a threshold function $\mathcal{T}(; R)$ that separates the low and high frequency components according
to the radius $R$. The formal definition of the equation $z_l,z_h=\mathcal{T}(z; R)$ is as follows:
\begin{equation} 
	\begin{split}
		\mathbf{z}_{l}(i, j)=\left\{\begin{array}{ll}\mathbf{z}(i, j), & \text { if } d\left((i, j),\left(c_{i}, c_{j}\right)\right) \leq R \\ 0, & \text { otherwise }\end{array}\right.,\\
		\mathbf{z}_{h}(i, j)=\left\{\begin{array}{ll}0, & \text { if } d\left((i, j),\left(c_{i}, c_{j}\right)\right) \leq R \\ \mathbf{z}(i, j), & \text { otherwise }\end{array}\right.,
		\label{Eq:5}
	\end{split}
\end{equation}
where $z(i,j)$ represents the value of $z$ at position index $(i, j)$ and $c_i,c_j$ indicate the index of the center point of the image. We use $d(\cdot,\cdot)$ as the Euclidean distance.

In this paper, frequency decomposition and frequency mixing of the real image $x_r$ and the generated image $x_g$ are defined as follows:
%\footnote{Since images have 3 channels, the below operations are performed separately on each channels}:
\begin{equation} 
\begin{split}
z_r=\mathcal{F}(x_r),\quad &z^l_r,z^h_r=\mathcal{T}(z_r;R), \\
z_g=\mathcal{F}(x_g),\quad &z^l_g,z^h_g=\mathcal{T}(z_g;R),\\
x^l_r=\mathcal{F}^{-1}(z^l_r),\quad & x^h_r=\mathcal{F}^{-1}(z^h_r),\\
x^r_r=\mathcal{F}^{-1}(z^l_r,\hat{z}^h_r),\quad & x^f_r=\mathcal{F}^{-1}(z^l_r,z^h_g),\\
x^l_g=\mathcal{F}^{-1}(z^l_g),\quad & x^h_g=\mathcal{F}^{-1}(z^h_g),\\
x^f_g=\mathcal{F}^{-1}(z^l_g,\hat{z}^h_g),\quad & x^r_g=\mathcal{F}^{-1}(z^l_g,z^h_r),\\
\label{Eq:4}
\end{split}
\end{equation}
where $\hat{z}^h_r$ represents the high-frequency components of another real image that is different from $x_r$. $\hat{z}^h_g$ represents the high-frequency components of another generated image that is different from $x_g$. 

The equations in Eq (\ref{Eq:4}) define the different frequency components of the image, some are frequency decompositions ($x^l_r$, $x^h_r$, $x^l_g$, and $x^h_g$) and some are frequency mixtures ($x^r_r$, $x^f_r$, $x^f_g$, and $x^r_g$). Properties of these frequency components will be discussed in the following sections.

%\subsection{Responses of the Discriminator to Different Frequency Components}
\subsection{High Frequency Filter (HFF) and High Frequency Confusion (HFC)}
\begin{figure*}
	\includegraphics[width=0.9\textwidth]{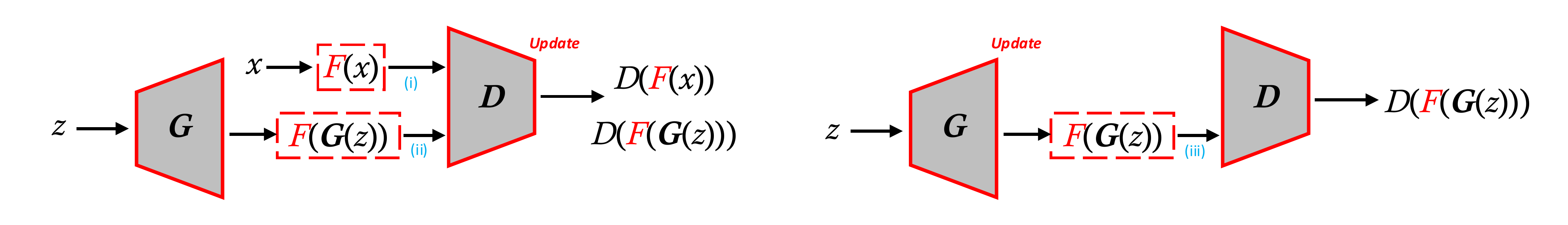}
	\centering
	\caption{Overview of the proposed frequency operations (HFF or HFC) for updating discriminator (left) and generator(right). }
	\label{figure:overview of different HFF}
\end{figure*}
\begin{figure}
	
	\includegraphics[width=0.5\textwidth]{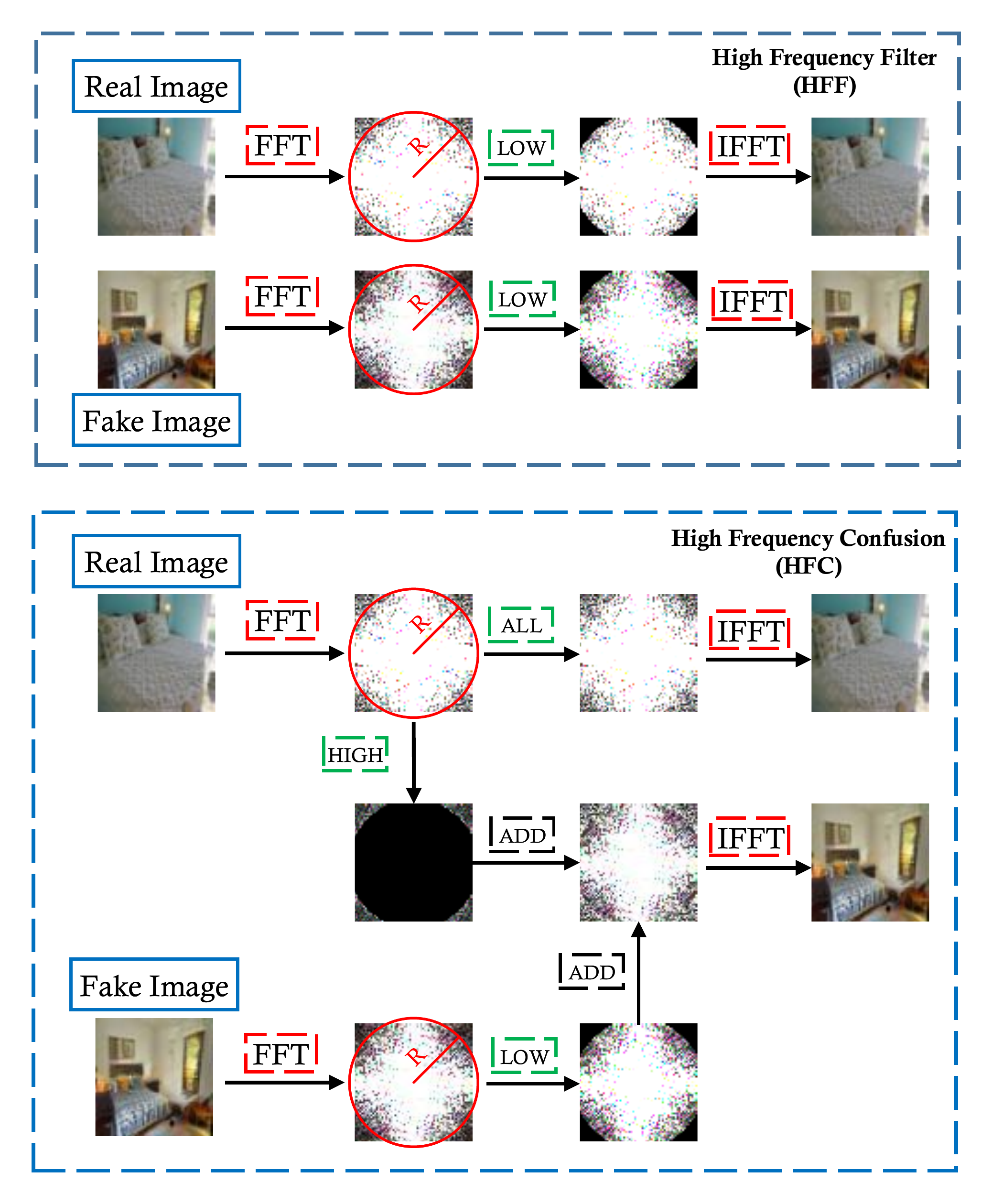}
	
	\centering
	\caption{The framework of the proposed methods High Frequency Filter (HFF) and High Frequency Confusion (HFC), in which Fast Fourier Transform (FFT) is a fast method to compute the Discrete Fourier Transform (DFT) of a sequence and IFFT is Inverse FFT. The operations LOW, HIGH and ALL are used to select the low-frequency components ($x^l$), high-frequency components ($x^h$) and all-frequency components ($x^l,x^h$) of the image ($x$).}
	\label{figure:5}
\end{figure}

In this subsection, we first analyze the response of discriminator to different frequency components and demonstrate that discriminator in GANs is sensitive to high-frequency components of the images. And then, we propose two frequency operations to eliminate the bias between the generated images and the real images during the training of GANs.

First, we analyze the responses of the discriminator to different frequency components introduced in Eq (\ref{Eq:4}). The mean values of the discriminator output on the CIFAR-10 dataset for different frequency components of real images are shown at the top part of Figure \ref{figure:response of the real images}, where $\text{real}$, $\text{real\_low}$, $\text{real\_real}$, $\text{real\_fake}$, and $\text{real\_high}$ represent $x_r$, $x^l_r$, $x^r_r$, $x^f_r$, and $x^h_r$, respectively. Furthermore, the visualization of different frequency components in the real images is also shown at the bottom part of Figure \ref{figure:response of the real images}. The result illustrates that the trained discriminator has differences in response to different frequency components ($x^l_r$, $x^r_r$, and $x^f_r$), even when $R\geq 16$. It is worth noting that \textbf{under different threshold radius (R), (1) the output of $\bm{x^f_r}$ is always lower than that of $\bm{x^r_r}$; (2) the output of $\bm{x^l_r}$ is always higher than that of $\bm{x^r_r}$}, which indicates that high-frequency components of images are critical to training the discriminator, and it also demonstrates that the bias between real images and generated images at high frequency is universal.

%The visualization illustrates that the high-frequency components are not visible to the human eyes when the frequency threshold radius (R) is greater than or equal to 16. Specifically,  $x_r$, $x^l_r$, $x^r_r$, and $x^f_r$ cannot be distinguished by humans when $R\geq 16$. However, the trained discriminator has differences in response to different frequency components, even when $R\geq 16$. It is worth noting that \textbf{under different threshold radius (R), the accuracy of $\bm{x^f_r}$ is always lower than that of $\bm{x^r_r}$} (for both training and testing sets), which indicates that high-frequency components of images are applied to train the discriminator of GANs, and it also demonstrates that the bias between real images and generated images at high frequency is universal and generalize.

Furthermore, we demonstrate the response of the discriminator to the generated images on the CIFAR-10 dataset in Figure \ref{figure:response of the fake images}, where $\text{fake}$, $\text{fake\_low}$, $\text{fake\_fake}$, $\text{fake\_real}$, and $\text{fake\_high}$ represent $x_g$, $x^l_g$, $x^f_g$, $x^r_g$, and $x^h_g$ in the Eq (4). Similar to the response of real images, \textbf{the output of $x^f_g$ is lower than that of $x^r_g$. The output of $x^l_g$ is always the highest, even higher than $x_g$, no matter what $R$ is.} 

Second, Figure \ref{figure:shortcut} illustrates some predicting defects of discriminator in the frequency domain. Original images and high-frequency components have the same prediction by the trained discriminator, while low-frequency components similar to the original image have a different prediction, which is not in line with human expectations. The results further demonstrate that discriminator of GANs is sensitive to high-frequency components of images.

%In summary, it is reasonable to believe that the discriminator in GANs is sensitive to high-frequency components that cannot be distinguished by humans and the high-frequency components of images affect the training of GANs.
It is reasonable to believe that the discriminator in GANs is sensitive to high-frequency components that is hard to be distinguished by humans. Meanwhile, the high-frequency components of images affect the training of GANs. Based on above analyses, we propose two frequency operations to eliminate high-frequency involvement during the training of GANs, which can improve the fit of the overall distribution, especially low-frequency components. Figure \ref{figure:overview of different HFF} illustrates the overview of the HFF and HFC for updating discriminator and generator during GANs training, the loss functions are:
\begin{equation} 
\begin{split}
 \min _{\phi} \max _{\theta} f(\phi, \theta) &=\mathbb{E}_{x \sim p_{r}}\left[g_{1}\left(D_{\theta}({F}(x))\right)\right] \\ &+\mathbb{E}_{z \sim p_{z}}\left[g_{2}\left(D_{\theta}\left({F}(G_{\phi}(z))\right)\right)\right], 
\label{Eq:6}
\end{split}
\end{equation}
where $F$ is required to be the same function (HFF or HFC) and the same threshold radius (R) across the three places illustrated in Figure \ref{figure:overview of different HFF}. The framework of HFF and HFC can be seen in Figure \ref{figure:5}, where HFF uses a filter to filter out the high-frequency components of the images, while HFC mixes two types of images by replacing the high-frequency components of the generated images with the high-frequency components of the real images. The form can be expressed as:
\begin{equation} 
\begin{split}
\begin{aligned} 
HFF(x_r,x_g;R)&=x^l_r,x^l_g\\
HFC(x_r,x_g;R)&=x_r,(x^l_g,x^h_r)
\end{aligned}
\label{Eq:7}
\end{split}
\end{equation}

\begin{table*}
	\caption{FID with using HFF in different places for training the GANs on CIFAR-10 dataset. "Real only" applies HFF to real (i) (see Figure \ref{figure:overview of different HFF}); "Fake only" applies HFF to fake (ii); "Dis only"  applies HFF to both real (i) and fake (ii), but not G (iii); "All" applies HFF to real (i), fake (ii), and G (iii). All FID scores are calculated on 50K training samples.
		\label{Tab:dif_HFF}}
	\centering
	\begin{tabular}{ccccccccccc}
		\toprule
		\multirow{2}{*}{Method}&\multicolumn{3}{c}{Where $F$? }&\multicolumn{6}{c}{Radius (R)}\\ 
		\cline{2-4} 	\cline{6-11}
		&
		\makecell*[c]{(i)}&(ii)&(iii)
		&&12&14&16&18&20&22\\
		
		\midrule
		SNGAN(Baseline)&&&&&20.05&20.05&20.05&20.05&20.05&20.05\\
		\midrule
		SNGAN-HFF (Real only)&\checkmark&&&&26.65&23.59&20.02&20.3&19.68&20.16\\
		SNGAN-HFF (Fake only)&&\checkmark&&&350&350&100&68&20.07&20.67\\
		SNGAN-HFF (Dis only)&\checkmark&\checkmark&&&350&52.2&20.07&20.59&19.89&19.53\\
		SNGAN-HFF (All)&\checkmark&\checkmark&\checkmark&&21.29&21.06&18.74&18.5&18.54&19.59\\
		\bottomrule
	\end{tabular}
\end{table*}

\begin{figure}
	
	\includegraphics[width=0.5\textwidth]{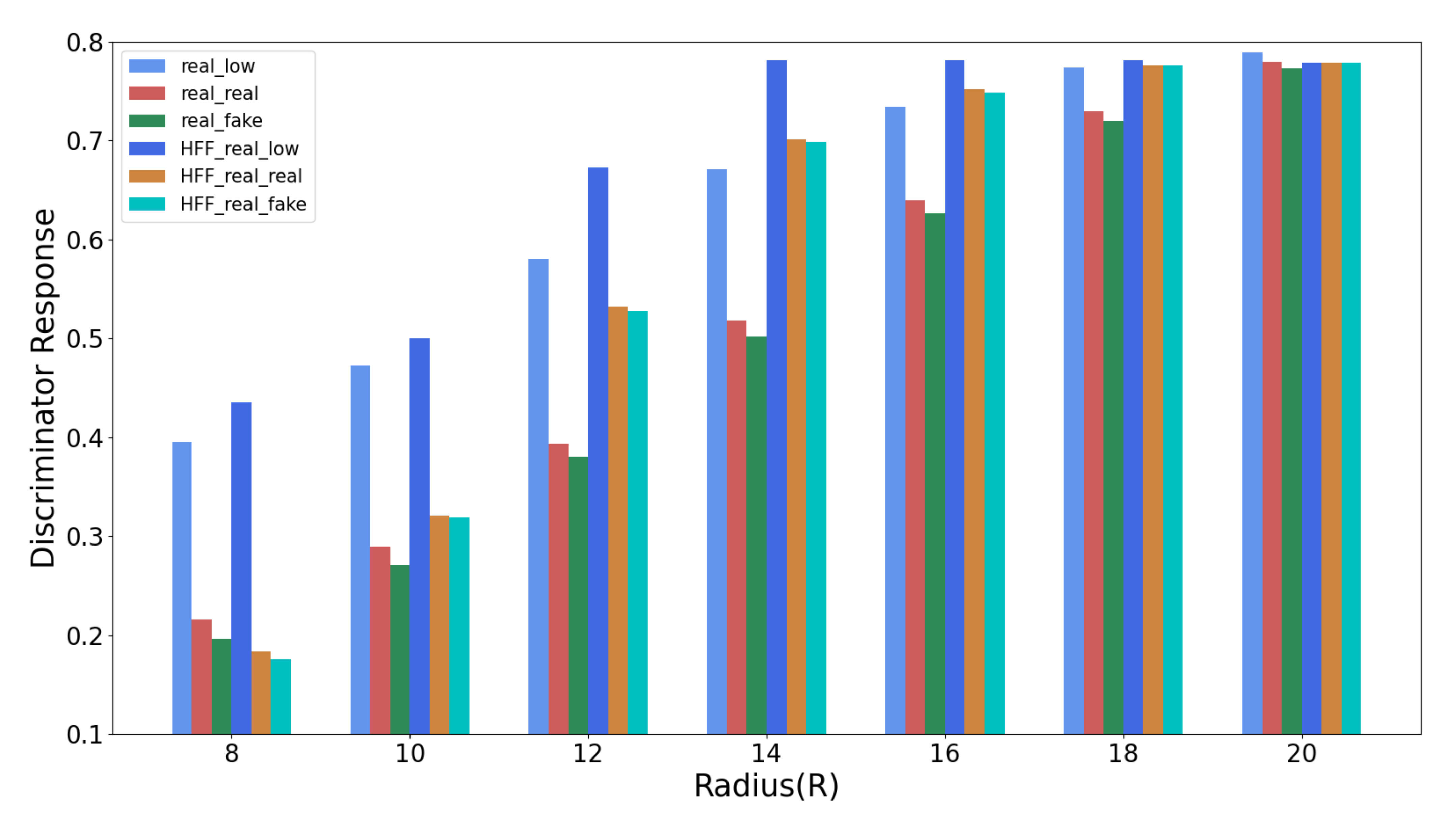}
	
	\centering
	\caption{The response of the discriminator to the different components of real images on the CIFAR-10 dataset, where $\text{real\_low}$, $\text{real\_real}$, and $\text{real\_fake}$ represent the output of discriminator trained by conventional pipeline to $x^l_r$, $x^r_r$, and $x^f_r$, respectively. Similarly, $\text{HFF\_real\_low}$, $\text{HFF\_real\_real}$, and $\text{HFF\_real\_fake}$ represent the output of discriminator with HFF-trained to $x^l_r$, $x^r_r$, and $x^f_r$, respectively. The other settings are the same as Figure \ref{figure:response of the real images}.}
	\label{figure:HFF_analusis}
\end{figure}

At the end, we also analyze the impact of the HFF for discriminator response. Figure \ref{figure:HFF_analusis} demonstrates the output of the discriminator to real images using the HFF before-and-after, respectively, where HFF indicates that the discriminator is trained with the proposed HFF method. The results illustrate: \textbf{(1) the discriminator output of $\text{HFF\_real\_low}$ is almost higher (i.e. more confidence)  than that of $\text{real\_low}$; (2) The gap between $\text{HFF\_real\_real}$ and $\text{HFF\_real\_fake}$ is less than the gap between $\text{real\_real}$ and $\text{real\_fake}$. In particular, when the filter Radius (R) is greater than 16, the gap between $\text{HFF\_real\_real}$ and $\text{HFF\_real\_fake}$ is 0; (3) When the radius is greater than 8, the discriminator outputs of $\text{HFF\_real\_real}$ and $\text{HFF\_real\_fake}$ are also greater than that of $\text{real\_real}$ and $\text{real\_fake}$.} These three points demonstrate that the proposed HFF alleviates the sensitivity of discriminator to high-frequency and improves the fit of discriminator to the low-frequency components.
\section{Experiments}
In the experimental section, we evaluate the proposed methods (HFF and HFC) on GANs training of different loss functions, network architectures, and datasets. The results demonstrate that high-frequency bias affects the fit of generator, and the proposed methods improve the generative performance of GANs. The details of datasets and experiment settings are shown in Section A of the \textbf{Supplementary Materials.}

\begin{figure}
	
	\includegraphics[width=0.5\textwidth]{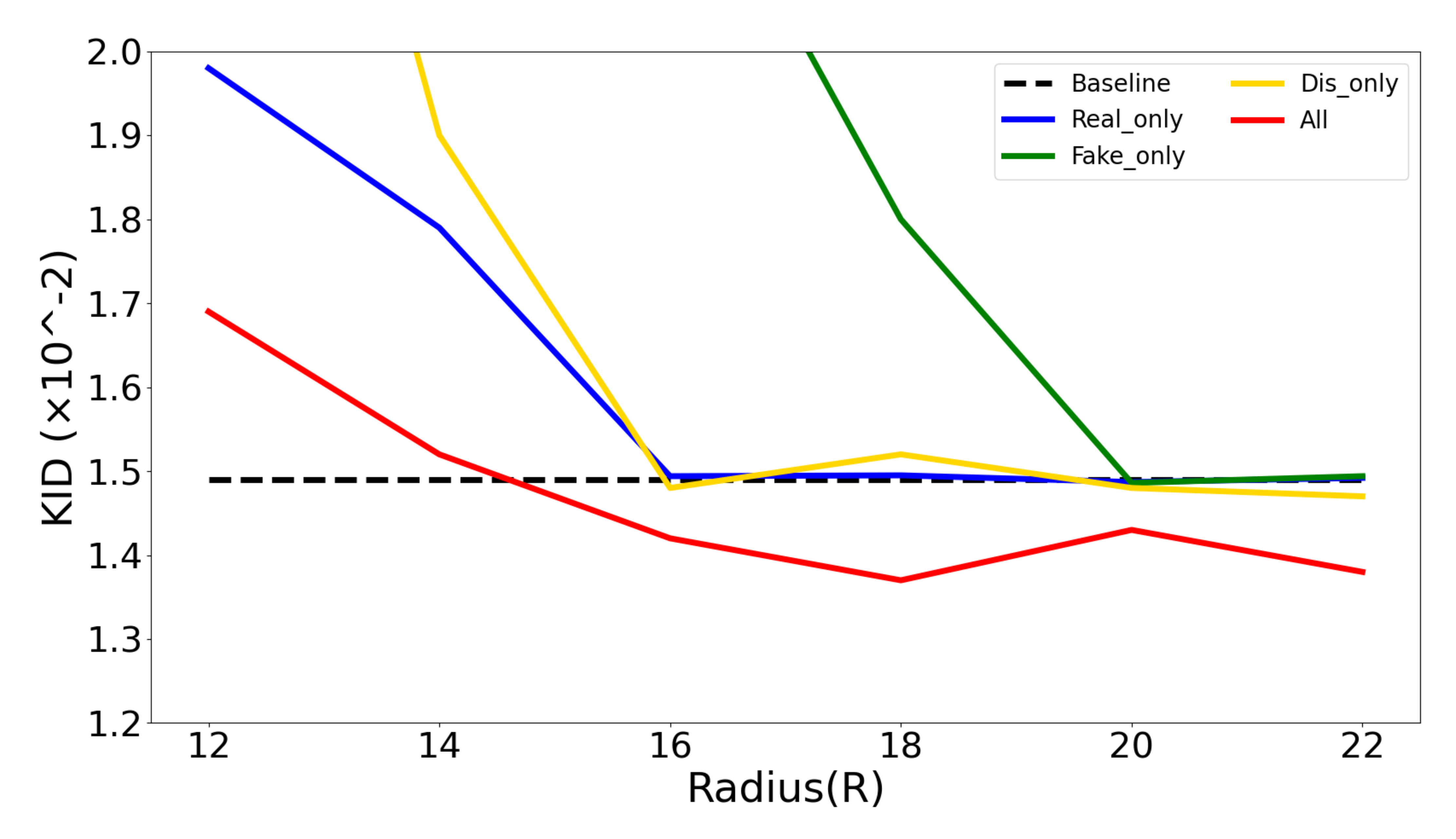}
	
	\centering
	\caption{KID with using HFF in different places for training the SNGAN on the CIFAR-10 dataset. All KID scores are calculated on 50K testing samples. The other settings are the same as in Table \ref{Tab:dif_HFF}.}
	\label{figure:6}
\end{figure}

\subsection{The Results on CIFAR-10 Dataset with Different Places}
In this subsection, we apply the proposed HFF to different places ((i), (ii), (iii) in Figure \ref{figure:overview of different HFF}) of the GANs training. Frechet Inception Distance (FID) and Kernel Inception Distance (KID) with using HFF in different places for training the GANs on the CIFAR-10 dataset are shown in Table \ref{Tab:dif_HFF} and Figure \ref{figure:6}, respectively. The results illustrate: (1) Excessive filtering of high-frequency components makes the training of "Real only", "Fake only", and "Dis only" methods fail, and reduces the performance of the "All" method; (2) When the threshold radius (R) is relatively large, performance of "Real only", "Fake only", and "Dis only" methods are similar to that of 
existing GANs (baseline), while performance of the "All" method is better than that of baseline. These results show that adding artificial high-frequency bias (using "Real only" or "Fake only" methods only filter high-frequency components of real or fake images, which introduces the artificial high-frequency bias during the GANs training) between the real images and the generated images has little effect on the training of GANs, while removing the high-frequency bias between the generated images and the real images ("All") improves the performance of GANs, which indicates that both natural and artificial high-frequency biases are not beneficial to the training of GANs.

\subsection{The Results on CIFAR-10 Dataset with Different Loss Functions}
\begin{figure}
	
	\includegraphics[width=0.5\textwidth]{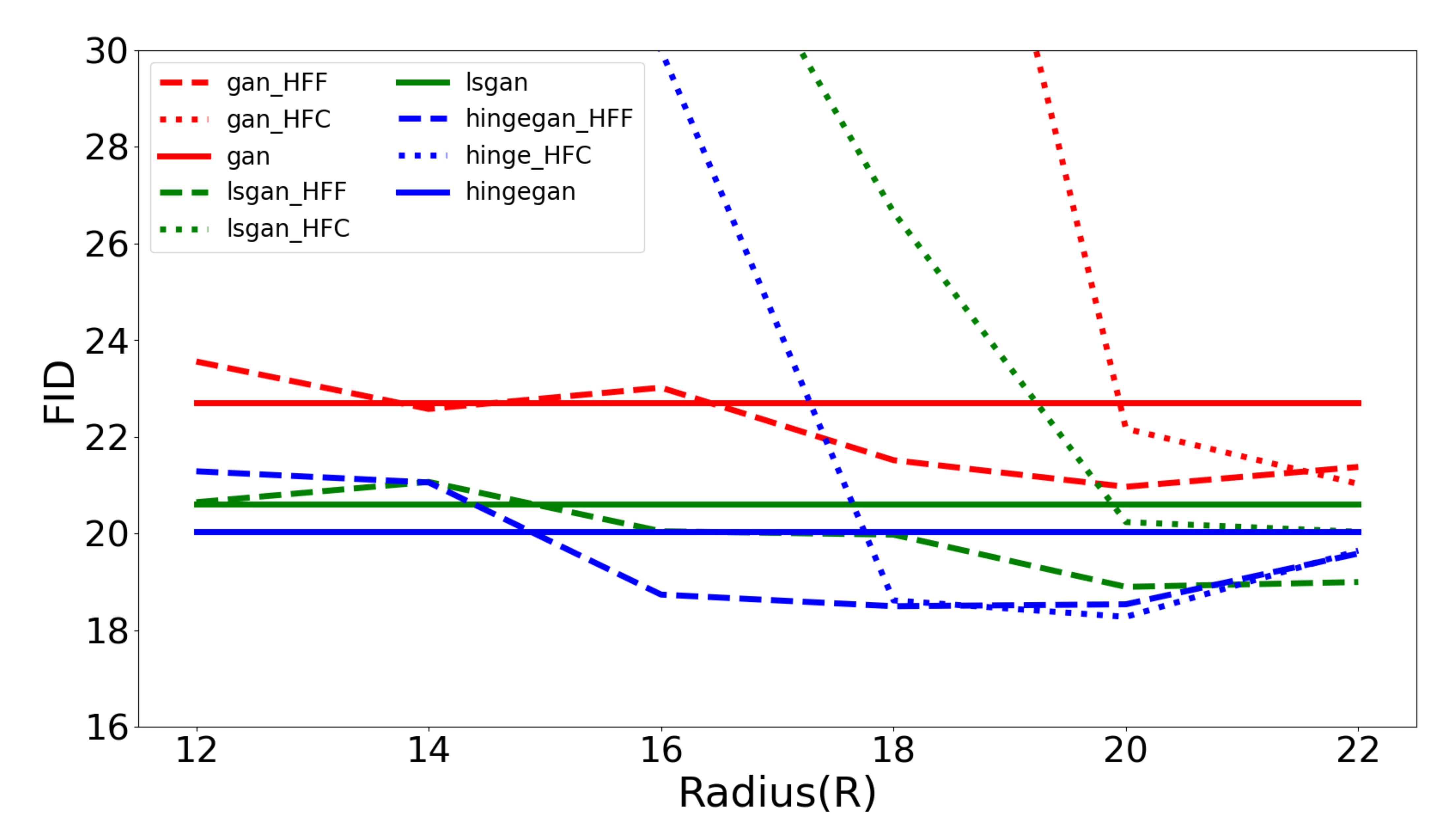}
	
	\centering
	\caption{FID on CIFAR-10 dataset with different loss functions in SNGAN. All FID scores are calculated on 50K training samples.}
	\label{figure:7}
\end{figure}

This subsection validates whether our proposed frequency opeartions are robust to different loss functions. Three different loss functions (vanilla GAN \cite{goodfellow2014generative}, LSGAN \cite{mao2017least}, hinge GAN \cite{lim2017geometric}) are used to train SNGAN on the CIFAR-10 dataset. FID and Inception Score (IS) are shown in Figure \ref{figure:7} and Table \ref{Tab:loss function}, respectively. Compared to HFF, it is easy to understand that HFC fails to train at small Radius. HFC replaces the high-frequency components of the generated images with the high-frequency components of the real images, which leads to a significant discontinuity in the spectrum domain of the generated images as shown in Figure \ref{figure:5}. The spectral discontinuity of the generated images leads to deviations between them and the real images, which leads to the fail of training. When R is large, this spectral discontinuity is minor, and the deviations between real images and generated images are imperceptible. Hence HFC is only suitable for larger radius. 

The proposed HFC and HFF both improve the performance of the baseline. Furthermore, the best performances of HFF and HFC are achieved when R equals to 20 (the vast majority of cases), which corresponds to the phenomenon illustrated in the upper and left part of the Figure \ref{figure:frequency_difference_all} (High-frequency bias occurs at the radius of 20 on the CIFAR-10 dataset). 
\begin{table*}
	\caption{Inception Score (IS) on CIFAR-10 dataset with different loss functions in SNGAN. All Inception scores are calculated on 50K training samples}
	\label{Tab:loss function}
	\centering
	\scalebox{1.1}{
		\begin{tabular}{ccccccccc}
			\toprule
			\multirow{2}{*}{Loss}&\multirow{2}{*}{Method}&\multicolumn{7}{c}{Radius (R)}\\ 
			
			\cline{3-9}
			&&\makecell*[c]{12}&14&16&18&20&22&$\infty$(Baseline)\\
			\midrule
			\multirow{2}{*}{gan}
			&HFF&$7.445^{\pm 0.15}$&$7.56^{\pm 0.1}$&$7.58^{\pm 0.04}$&$7.66^{\pm 0.05}$&$\textbf{7.79}^{\pm 0.08}$&$7.78^{\pm 0.04}$&\multirow{2}{*}{$7.68^{\pm 0.09}$}\\
			&HFC&$3.2^{\pm 0.02}$&$3.5^{\pm 0.01}$&$4.5^{\pm 0.05}$&$6.36^{\pm 0.11}$&{$7.42^{\pm 0.05}$}&$\textbf{7.76}^{\pm 0.09}$&\\
			\multirow{2}{*}{lsgan}
			&HFF&$7.63^{\pm 0.05}$&$7.76^{\pm 0.08}$&$7.76^{\pm 0.12}$&$7.66^{\pm 0.09}$&$\textbf{7.89}^{\pm 0.1}$&$7.89^{\pm 0.05}$&\multirow{2}{*}{$7.74^{\pm 0.14}$}\\
			&HFC&$3.2^{\pm 0.03}$&$5.51^{\pm 0.06}$&$6.22^{\pm 0.12}$&$7.47^{\pm 0.05}$&{$\textbf{7.82}^{\pm 0.02}$}&${7.74}^{\pm 0.19}$&\\
			\multirow{2}{*}{hinge gan}
			&HFF&$7.79^{\pm 0.09}$&$7.77^{\pm 0.03}$&$7.86^{\pm 0.1}$&$7.85^{\pm 0.08}$&$\textbf{7.95}^{\pm 0.07}$&$7.86^{\pm 0.12}$&\multirow{2}{*}{$7.84^{\pm 0.08}$}\\
			&HFC&$3.5^{\pm 0.04}$&$4.3^{\pm 0.04}$&$5.67^{\pm 0.14}$&$7.89^{\pm 0.03}$&{$\textbf{8.01}^{\pm 0.06}$}&${7.9}^{\pm 0.12}$&\\
			\bottomrule
	\end{tabular}}
\end{table*}

\subsection{Evaluation on Different Datasets and Different Architectures}
In this subsection, HFF and HFC are adopted on some popular models, such as SNGAN, SSGAN, and InfoMAXGAN. Hinge loss is selected for all experiments. The baseline codes are available on Github\footnote{\url{https://github.com/kwotsin/mimicry}}. We evaluate our methods on five different datasets at multiple resolutions: CIFAR-10 ($32\times 32$) \cite{krizhevsky2009learning}, CIFAR-100 ($32\times 32$) \cite{krizhevsky2009learning}, STL-10 ($48\times48$) \cite{coates2011analysis}, LSUN-Bedroom ($64 \times64$) \cite{yu2015lsun}, and CelebA ($128 \times128$). Also, three metrics are adopted to evaluate the quality of generated images: Fréchet Inception Distance (FID) \cite{heusel2017gans}, Kernel Inception Distance (KID) \cite{binkowski2018demystifying}, and Inception Score (IS) \cite{salimans2016improved}. 
\begin{figure}
	\begin{subfigure}[b]{0.23\textwidth}
		\centering
		\includegraphics[width=\textwidth]{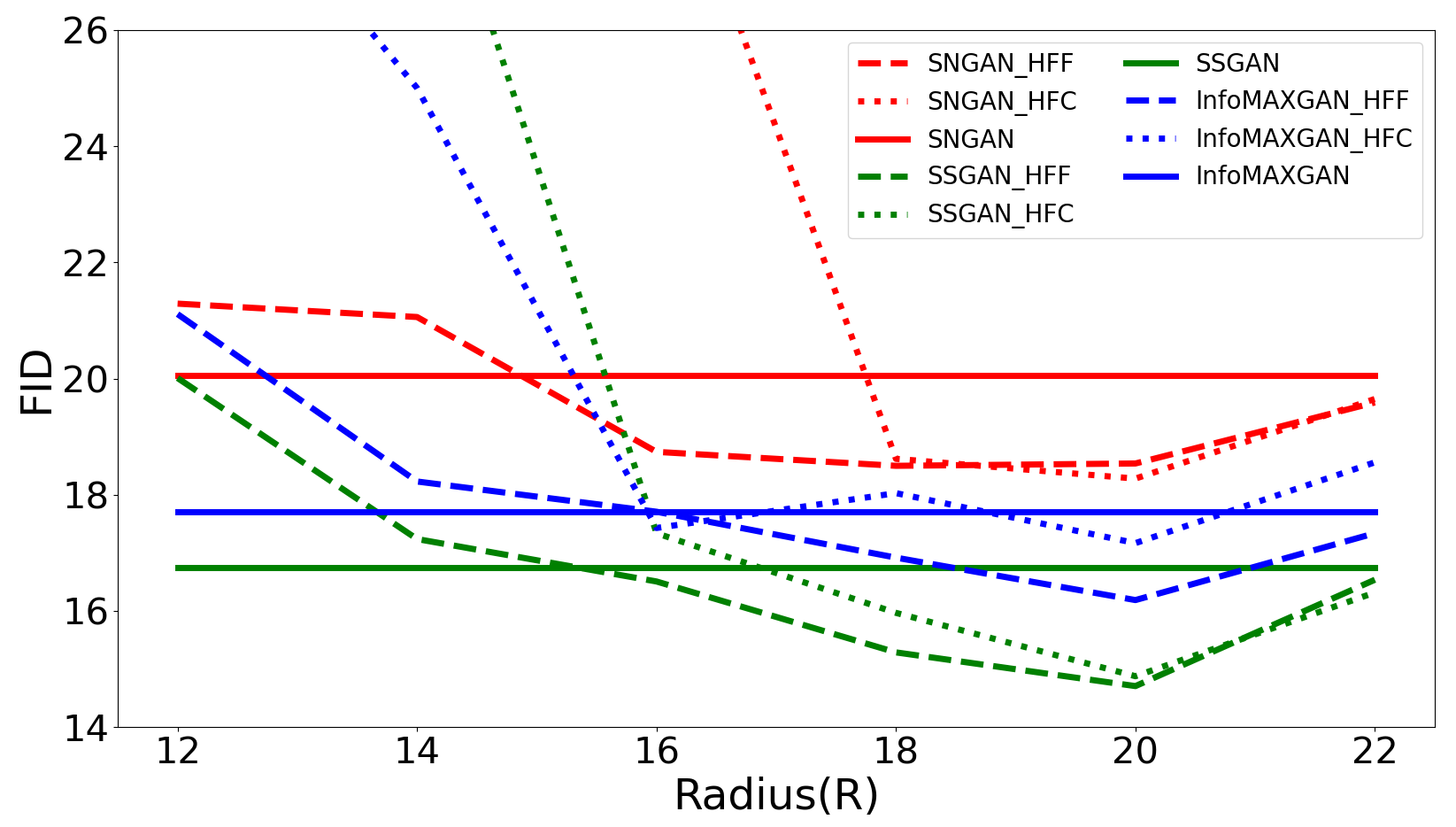}
		\caption{CIFAR-10}
		\label{figure:adv attack without adv training}
	\end{subfigure}
	\begin{subfigure}[b]{0.23\textwidth}
		\centering
		\includegraphics[width=\textwidth]{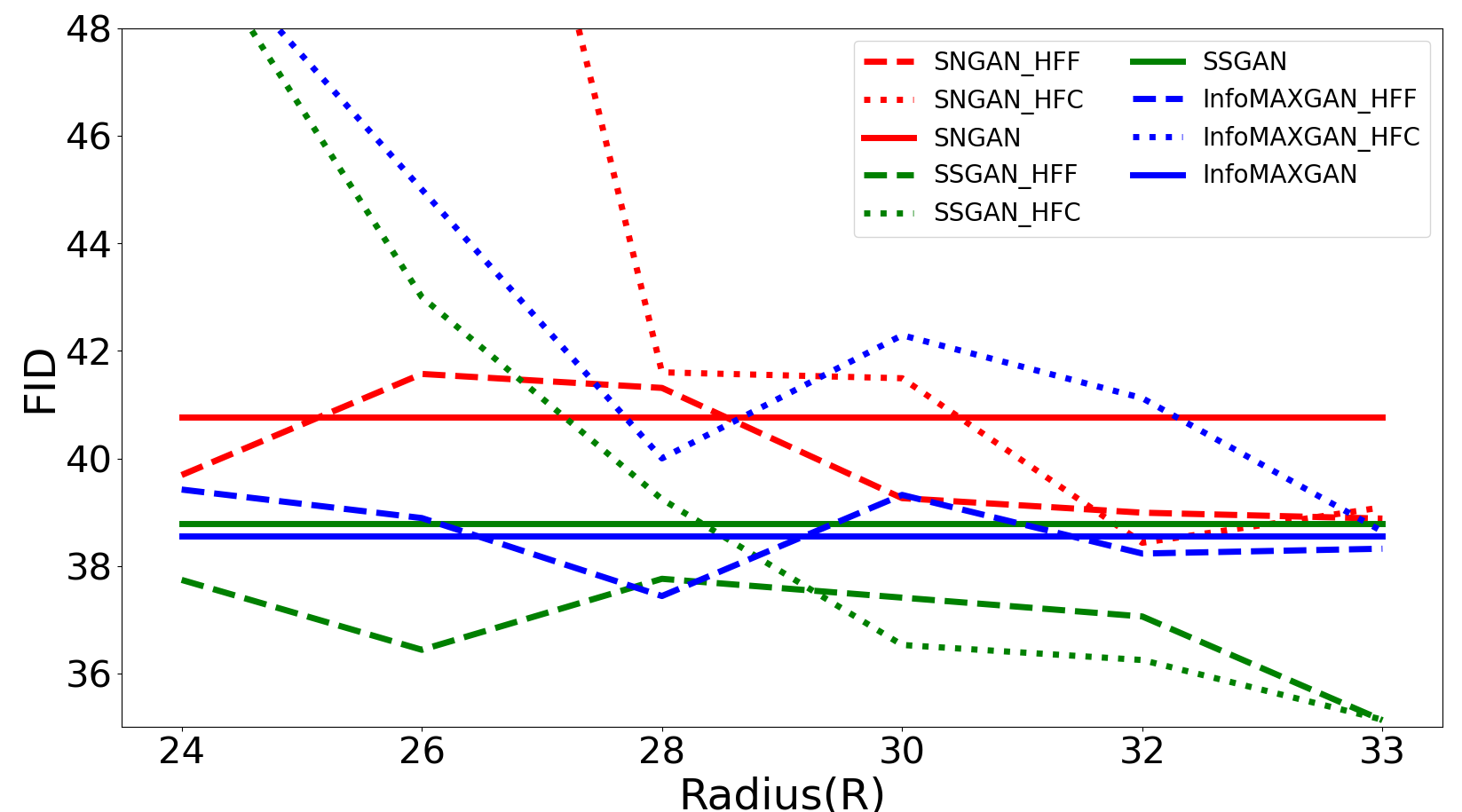}
		\caption{STL-10}
		\label{figure:adv attack with adv training}
	\end{subfigure}
	\begin{subfigure}[b]{0.23\textwidth}
		\centering
		\includegraphics[width=\textwidth]{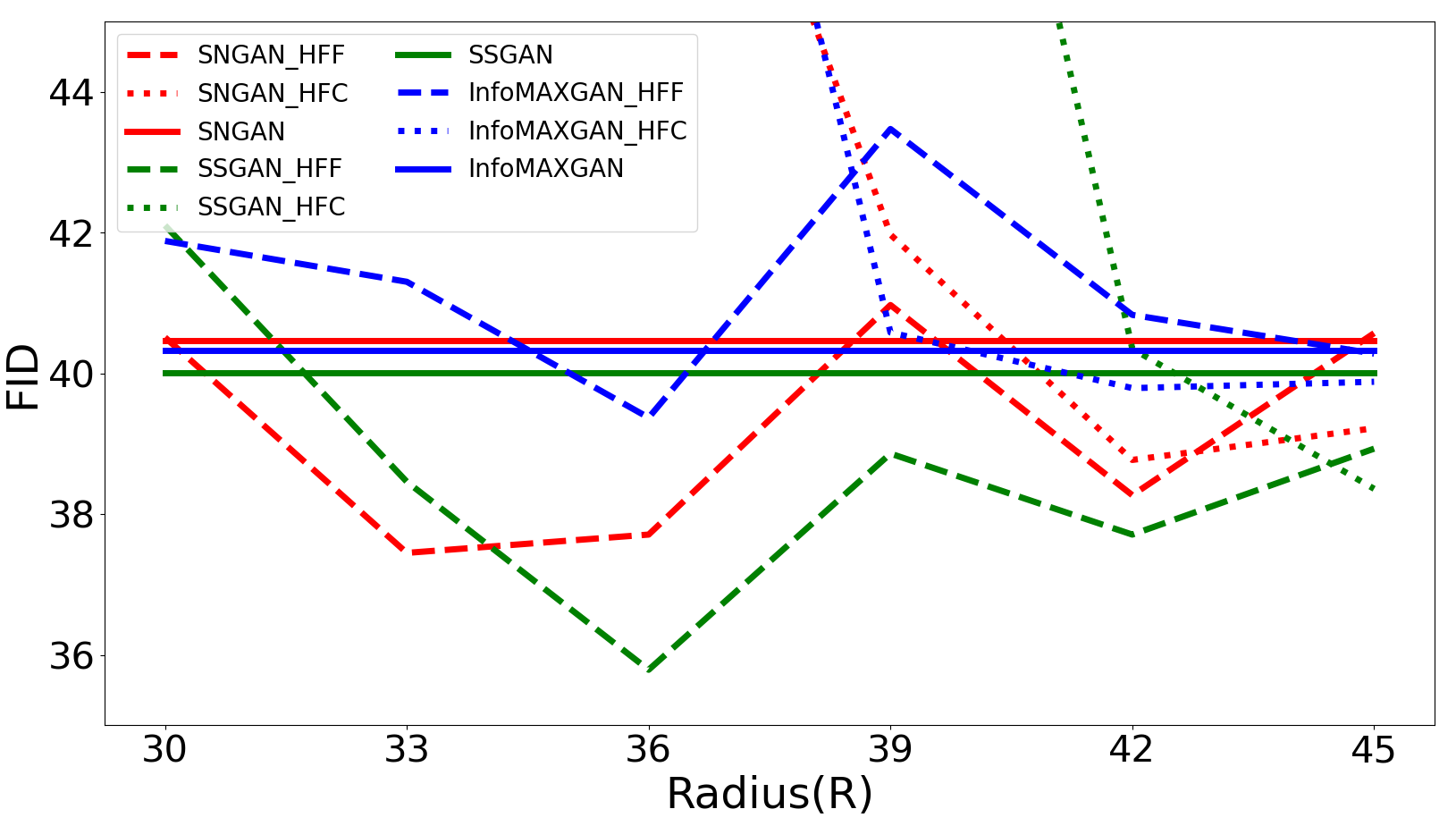}
		\caption{LSUN-Bedroom}
		\label{figure:adv attack with adv training}
	\end{subfigure}
\begin{subfigure}[b]{0.23\textwidth}
	\centering
	\includegraphics[width=\textwidth]{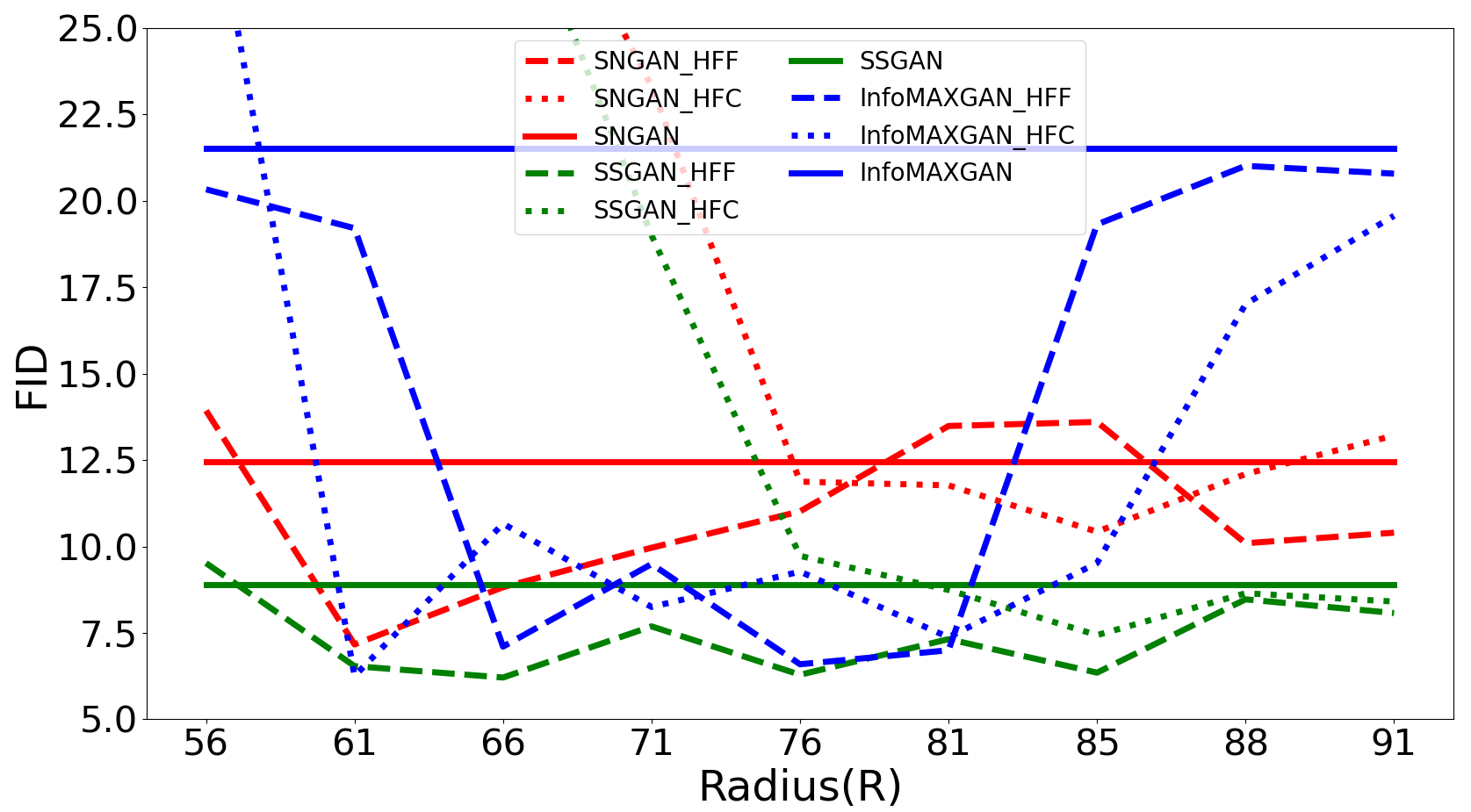}
	\caption{CelebA}
	\label{figure:adv attack without adv training}
\end{subfigure}
	\centering
	\caption{FID results for different Radius (R) on different datasets}
	\label{figure:FID}
\end{figure}
\begin{table*}
	\caption{FID, KID, and IS scores of popular models on different datasets. FID and KID: lower is better. IS: higher is better.\label{tab:SOTA}}
	\centering
	\scalebox{1}{
		\begin{tabular}{ccccc|ccc|ccc}
			\toprule
			\multirow{3}{*}{\makecell*[c]{Metric}} &\multirow{3}{*}{Dataset} &\multicolumn{9}{c}{\makecell*[c]{Models}} \\ 
			\cline{3-11} 
			\specialrule{0em}{2pt}{2pt}
			&&
			\multicolumn{3}{c}{\makecell*[c]{SNGAN}}&\multicolumn{3}{c}{\makecell*[c]{SSGAN}}&\multicolumn{3}{c}{\makecell*[c]{InfoMAX-GAN}}\\
			
			&&None&HFF&HFC&None&HFF&HFC&None&HFF&HFC\\
			\midrule
			\multirow{5}{*}{FID $(\downarrow)$}&{CelebA}&12.46&7.17&10.43&8.90&\textbf{6.21}&7.44&21.52&6.59&6.25\\
			&{LSUN-Bedroom}&40.47&37.71&38.77&40.01&\textbf{35.79}&38.36&40.32&39.37&39.79\\
			
			&{STL-10}&40.77&38.99&38.43&38.78&\textbf{35.13}&35.14&38.55&37.44&38.64\\
			
			&{CIFAR-100}&23.70&21.45&21.98&22.43&20.27&21.51&21.36&20.41&\textbf{19.16}\\
			
			&{CIFAR-10}&20.05&18.54&18.28&16.75&\textbf{14.71}&14.88&17.71&16.19&17.17\\
			\cline{1-11} 
			\specialrule{0em}{2pt}{2pt}
			\multirow{5}{*}{\makecell[c]{KID $(\downarrow)$ \\($\times 10^{-2}$)}}&{CelebA}&0.91&0.51&0.67&0.62&\textbf{0.42}&0.51&1.65&0.43&0.43\\

			&{LSUN-Bedroom}&4.86&\textbf{4.22}&4.65&4.75&4.36&4.53&4.81&4.66&4.72\\
			&{STL-10}&3.84&3.68&3.77&3.62&\textbf{3.28}&3.31&3.69&3.54&3.72\\
			&{CIFAR-100}&1.58&\textbf{1.31}&1.35&1.55&1.4&1.56&1.52&1.39&1.33\\
			&{CIFAR-10}&1.49&1.38&1.35&1.25&1.13&1.13&1.34&\textbf{1.12}&1.24\\
			\cline{1-11} 
			\specialrule{0em}{2pt}{2pt}
			\multirow{5}{*}{IS $(\uparrow)$}&CelebA&-&-&-&-&-&-&-&-&-\\
			&LSUN-Bedroom&-&-&-&-&-&-&-&-&-\\
			&STL-10&8.43&8.58&8.54&8.58&\textbf{8.79}&8.76&8.63&8.74&8.69\\
			&
			CIFAR-100&7.66&7.85&7.84&7.74&7.85&7.96&8.02&8.02&\textbf{8.1}\\
			&
			CIFAR-10&7.84&7.95&8.01&\textbf{8.13}&8.1&8.09&8.01&8.1&8.0\\
			\bottomrule
	\end{tabular}}
\end{table*}

\begin{figure}
	
 	\includegraphics[width=0.5\textwidth]{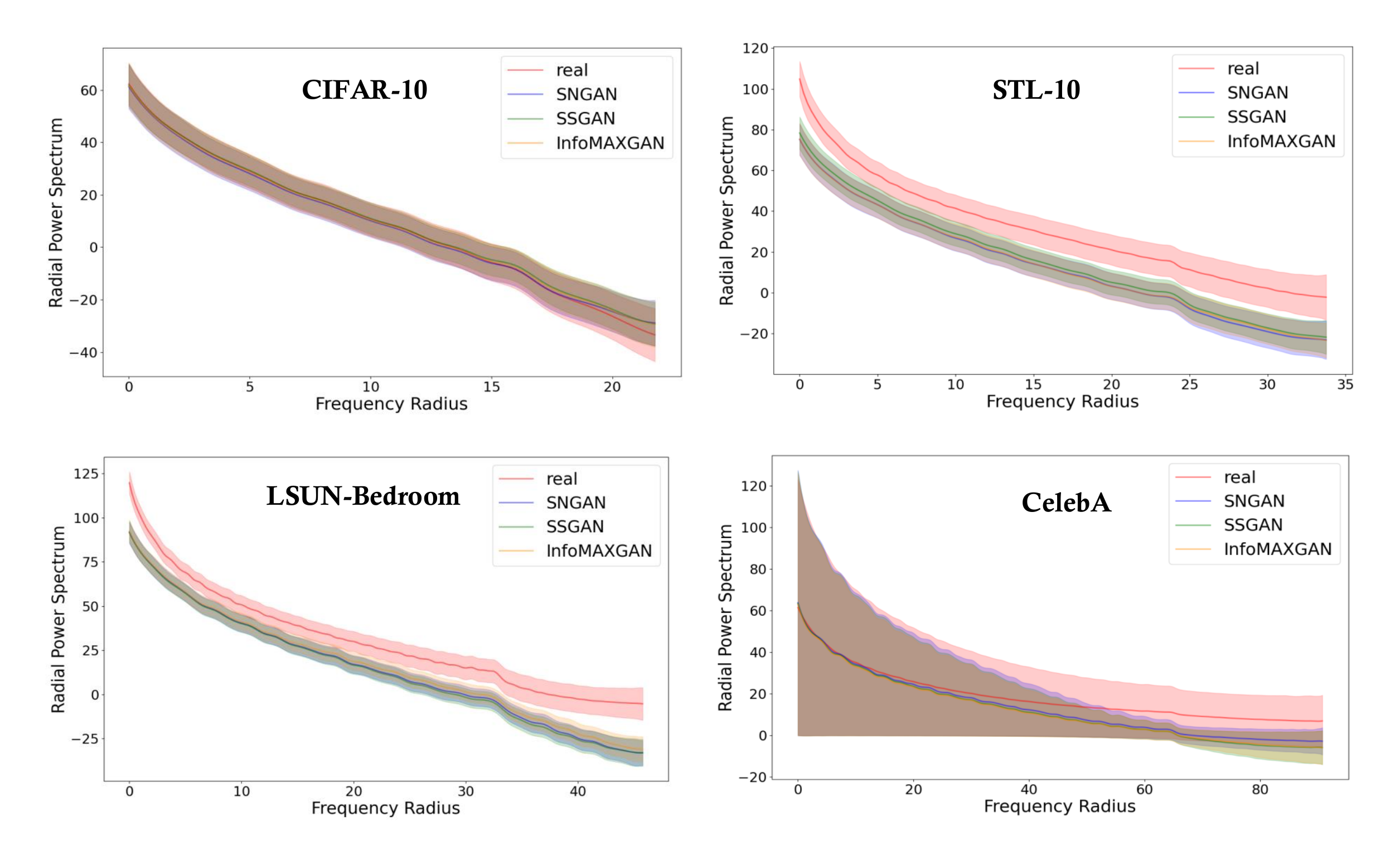}
	
	\centering
	\caption{The bias of Radial Power Spectrum between real and generated images on CIFAR-10, STL-10, LSUN-Bedroom, and CelebA datasets.}
	\label{figure:frequency_difference_all}
\end{figure}

Radius (R) is an important parameter to control the range of frequency during GANs training. Here, different R settings are evaluated on four datasets\footnote{Both CIFAR-10 and CIFAR-100 have the same resolution, so we only demonstrate the results on CIFAR-10, STL-10, LSUN-Bedroom, and CelebA datasets.} to investigate their effect on the training of GANs. First, the bias of Radial Power Spectrum between real and generated images on four datasets are demonstrated in Figure \ref{figure:frequency_difference_all}. Figure \ref{figure:frequency_difference_all} illustrates that the bias between real and generated images on a small dataset (CIFAR-10) mainly exists in the high-frequency components. But for big datasets, such as STL-10, LSUN-Bedroom, and CelebA, all-frequency components exist the bias between real and generated images, which introduces uncertainty into the setting of the optimal Radius (R). To investigate the effect of different settings of Radius (R), Figure \ref{figure:FID} illustrates the FID of different Radius on four datasets. It is clear that: the optimal Radius of HFF\footnote{The reason we mainly consider HFF is that HFC can cause model collapse when the Radius is small} is set at approximately the same location as the appearance of frequency bias (R=20) on the CIFAR-10 dataset. However, due to the underfitting on complex datasets (STL-10, LSUN-Bedroom, and CelebA), the frequency bias exists in the whole frequency domain. There is uncertainty in the optimal Radius for different models. 
\begin{table*}
	\caption{FID scores and training time of popular models on CIFAR-10, CIFAR-100, and STL-10 datasets, where the time is training time on SNGAN.
		\label{tab:SSD-GAN}}
	\centering
	\renewcommand\tabcolsep{5.0pt}
	\renewcommand\arraystretch{1.2}
	\begin{threeparttable}
		\begin{tabular}{cccccccccc}
			\toprule 
			\multirow{2}{*}{\makecell*[c]{Method}}  
			&\multicolumn{3}{c}{\makecell*[c]{CIFAR-10}}&\multicolumn{3}{c}{\makecell*[c]{CIFAR-100}}&\multicolumn{3}{c}{\makecell*[c]{STL-10}}\\
			\cmidrule(r){2-4} \cmidrule(r){5-7} \cmidrule(r){8-10}\\
			\specialrule{0em}{2pt}{2pt}
			&Time(h)&SNGAN&SSGAN&Time(h)&SNGAN&SSGAN&Time(h)&SNGAN&SSGAN\\
			\midrule
			None&18.1&20.05&16.75&18.4&23.70&22.43&36.0&40.77&38.78\\
			HFF&18.3&18.54&14.71&18.4&21.45&20.27&35.5&38.99&35.13\\
			HFC&18.2&18.28&14.88&18.3&21.98&21.51&36.2&38.43&35.14\\
			SSD&18.7&16.73&13.51&18.9&20.73&19.91&37.8&37.04&34.43\\
			%SSD+HFF(Both)&19.0&15.80&13.08&19.0&19.58&19.50&37.5&35.55&33.44\\
			%SSD+HFC(Both)&18.9&16.52&13.26&18.9&19.81&19.49&37.8&35.22&33.09\\
			SSD+HFF&19.0&\textbf{15.40}&\textbf{12.39}&21.1&\textbf{19.30}&\textbf{18.64}&40.3&35.03&\textbf{32.68}\\
			SSD+HFC&18.9&16.01&12.88&20.9&19.68&19.25&39.8&\textbf{34.43}&33.12\\
			\bottomrule
		\end{tabular}
	\end{threeparttable}
\end{table*}
\begin{comment}
\begin{table}
	\caption{FID between real images and real images with HFF or HFC on the CIFAR-10 dataset. }
	\label{Tab:FID_REAL_IMAGES}
	\centering
	
	\begin{tabular}{ccccccc}
		\toprule
		\multirow{2}{*}{Method}&\multicolumn{6}{c}{Radius (R)}\\ 
		
		\cline{2-7}
		&\makecell*[c]{14}&16&18&20&22&$\infty$(Baseline)\\
		\midrule
		HFF&90.34&48.07&5.713&5.28&5.21&5.21\\
		HFC&47.50&31.71&6.85&5.48&5.22&5.21\\
		\bottomrule
	\end{tabular}
\end{table}
\end{comment}
Furthermore, we also demonstrate the whole results under the optimal Radius (R) in Table \ref{tab:SOTA}. The results demonstrate that the proposed methods improve the performance of GANs on different architectures and datasets. For instance, compared to SNGAN, the improvement in FID of SNGAN-HFF is $7.5\%$, $9.5\%$, $4.4\%$, $6.8\%$, $42.5\%$ from CIFAR-10 (32*32), CIFAR-100 (32*32), STL-10 (48*48), LSUN (64*64) to CelebA (128*128). Similarly, compared to SSGAN, the improvement in FID of SSGAN-HFF is $12.2\%$, $9.6\%$, $9.4\%$, $10.6\%$, $30.2\%$ from CIFAR-10, CIFAR-100, STL-10, LSUN to CelebA. In particular, the improvement on the CelebA dataset is significant and impressive, which is open to interpretation. The motivation of our work is \textit{high-frequency difference between real and fake images reduces the fit of the generator to the low-frequency components, which is not beneficial for the GANs training.} Based on this hypothesis, it is reasonable to assume that the proposed method will realize better performance on large resolution datasets. Dzanic \textit{et al}. \cite{dzanic2019fourier} show that at higher resolutions, the differences in frequency domain between real and fake images are more easily observed. Similarly, Figure \ref{figure:frequency_difference_all} also illustrates this phenomenon.

%Specifically, the optimal Radius for both HFF and HFC is chosen to be 20 on the CIFAR-10, and CIFAR-100 datasets, the optimal Radius for HFF of SNGAN, SSGAN, and InfoMAXGAN on STL-10 dataset are chosen to be 32, 33, and 28 respectively, the optimal Radius for HFC of SNGAN, SSGAN, and InfoMAXGAN on STL-10 dataset are chosen to be 32, 33, and 33 respectively, the optimal Radius for HFF of all models on LSUN-Bedroom dataset are chosen to be 36, and the optimal Radius for HFC of SNGAN, SSGAN, and InfoMAXGAN on LSUN-Bedroom dataset are chosen to be 42, 45, and 42 respectively.

For qualitative comparisons, we present randomly sampled images generated by SSGAN and SSGAN with HFF for CIFAR-10, CIFAR-100, STL-10, LSUN-Bedroom, and CelebA datasets in Section B of \textbf{Supplementary Materials.}
\begin{figure}
	
	\includegraphics[width=0.5\textwidth]{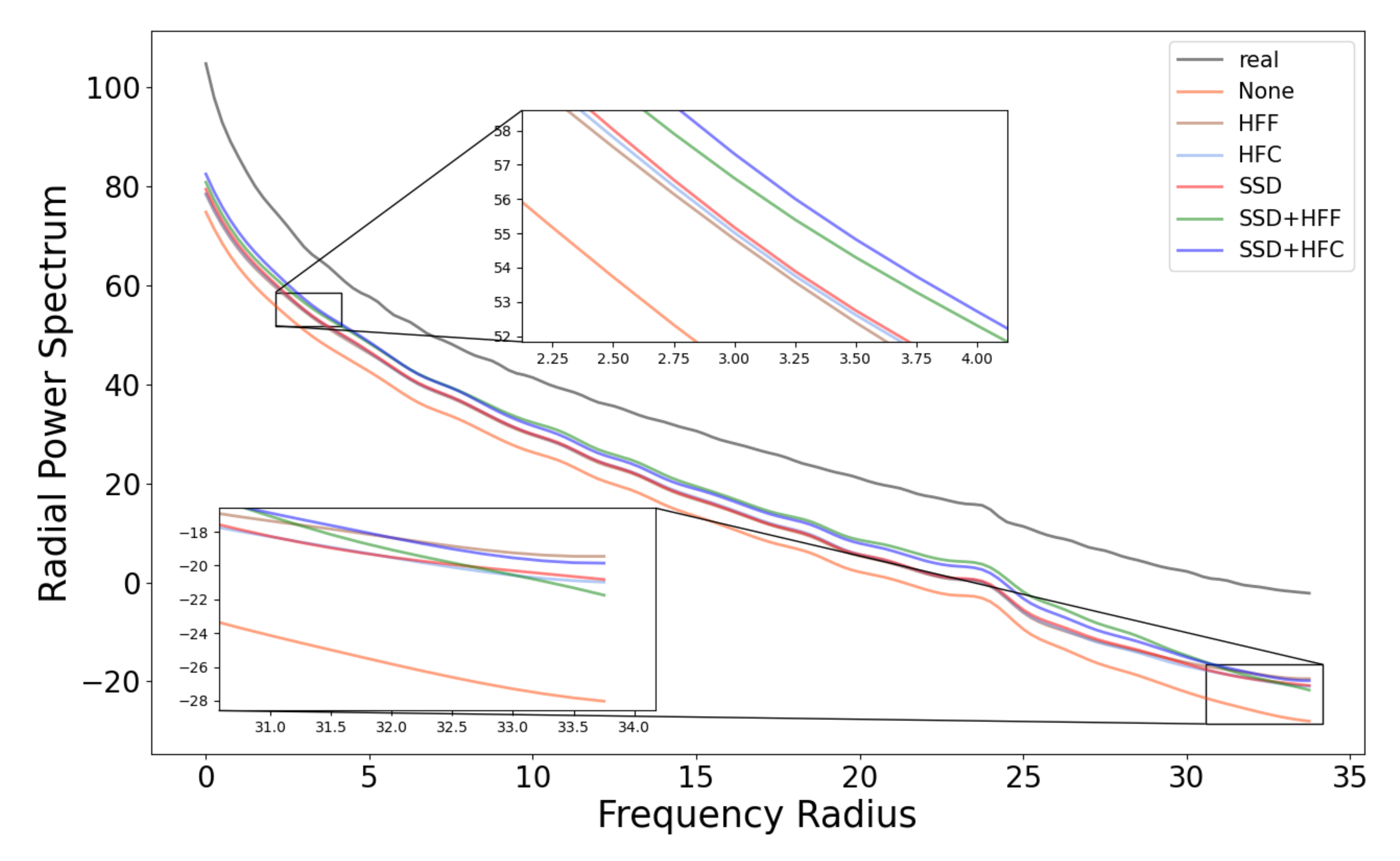}
	
	\centering
	\caption{The Radial Power Spectrum (RPS) of different SNGAN-based methods on STL-10 dataset by averaging 1K samples.}
	\label{figure:frequency_difference_stl10}
\end{figure}
\subsection{The Comparison with SSD-GAN}
SSD-GAN \cite{chen2021ssd} is also a method related to frequency domain in GANs training, which reveals that frequency missing in the discriminator results in a significant spectrum discrepancy between real and fake images. To mitigate this problem, SSD-GAN boosts a frequency-aware classifier into the discriminator to measure the realness of images in both spatial and spectral domains, which improves the spectral discrimination of discriminator. However, the frequency biases between real and fake images have not been eliminated. In our study, we hypothesize that high-frequency differences make the discriminator focus on high-frequency components excessively, which hinders generator from fitting the important low-frequency components. The results indicate that our methods can be combined with SSD-GAN, which further boosts the performance of GANs. 

Furthermore, we compare the proposed HFF and HFC with SSD-GAN on multiple datasets, such as CIFAR-10, CIFAR-100, and STL-10 datasets. FID scores and the training time on SNGAN are given in Table \ref{tab:SSD-GAN}, where we only use the HFF or HFC in spatial discriminator. The results indicate that the proposed methods improve the performance of GANs with little cost and also boost the performance of SSD-GAN. We also show the Radial Power Spectrum of different SNGAN-based methods on the STL-10 dataset in Figure \ref{figure:frequency_difference_stl10}. Compared to SNGAN (None), the low-frequency spectrum discrepancy between images generated by the proposed methods and real data is significantly reduced, where the method "SSD+HFC" achieves the best performance. Although the proposed method reduces the focus on the high-frequency component, this does not affect the high-frequency fit and even reduces the high-frequency bias. 

%Last but not the least, we analyze the effect of the frequency components on the FID. Table \ref{Tab:FID_REAL_IMAGES} demonstrates that the FID measurement is responsive to all frequency components. According to it, better fitting of all frequency components contributes to the improvement of the FID, and low-frequency components contribute more to the calculation of FID. We hypothesize that high-frequency components reduce the sensitivity of discriminator to low-frequency components, which prevents the fit of low-frequency components. When the low-frequency components are well fitted, the fit of the high-frequency components is the icing on the cake. Therefore, we only apply HFF or HFC methods in spatial discriminator. The use of HFF in the spectral discriminator affects the fit of the generator to high frequencies. 

\section{Conclusion and Outlook}
This paper analyzes the response of discriminator to different-frequency components and indicates that high-frequency bias prevents the fit of the overall distribution, especially the low-frequency components, which is not conducive to the training of GANs. To avoid the high-frequency differences between the generated images and the real images during the GANs training, two frequency operations (HFF and HFC) are introduced in this paper. We provide empirical evidence that the proposed HFF and HFC improve the generative performance with a little cost. 

Frequency analysis is an interesting and new perspective on neural networks. Recent works have shown that high-frequency components are related to the generalization and robustness of CNNs. This paper aims for the first pilot study of different frequency components on GANs training. Although significant conclusions and remarkable results have been obtained, the modeling process of GANs for different frequency components is still unclear. In the future, we will verify the frequency principles (DNNs often fit target functions from low to high frequencies during the training) of generators and discriminators, and also explore the relationship between different frequency components and robustness and generalization of GANs training. We hope our work set forth towards a new perspective: Frequency analysis in GANs training.
%-------------------------------------------------------------------------
\section*{Acknowledgment}

The work is partially supported by the National Natural Science Foundation of China under Grand No.U19B2044 and No.61836011. We are grateful to Chaoyue Wang and Jing Zhang for support of this work.

\bibliographystyle{IEEEtran}
\bibliography{egbib}

\end{document}